\pdfoutput=1
\documentclass[11pt]{article}

\usepackage{acl}

\usepackage{times}
\usepackage{latexsym}

\usepackage[T1]{fontenc}
\usepackage[utf8]{inputenc}
\usepackage[colorinlistoftodos]{todonotes}
\usepackage{xcolor}
\usepackage{soul}

\usepackage{microtype}

\usepackage{inconsolata}

\usepackage{xcolor,colortbl} % for colors
\usepackage{amsmath} % for math symbols
\usepackage{amssymb} % for math symbols
\usepackage{wasysym} % for math symbols
\usepackage{graphicx}
\usepackage{multirow}
\usepackage{tikz,pgfplots}
\usepackage{booktabs}
\usepackage{comment}

\newcommand{\commentout}[1]{}

\makeatletter
\def\blfootnote{\xdef\@thefnmark{}\@footnotetext}
\makeatother

\newcommand{\authorspace}{\hspace{9pt}}

\title{Attributed Question Answering: \\ 
Evaluation and Modeling for Attributed Large Language Models}

\author{Bernd Bohnet$^*$ \authorspace Vinh Q. Tran$^*$ \authorspace Pat Verga$^*$ \\
\AND
\bf Roee Aharoni \authorspace Daniel Andor \authorspace Livio Baldini Soares \authorspace Massimiliano Ciaramita \\ \bf Jacob Eisenstein \authorspace Kuzman Ganchev \authorspace Jonathan Herzig  \authorspace Kai Hui \authorspace \\
\bf Tom Kwiatkowski \authorspace Ji Ma \authorspace Jianmo Ni \authorspace Lierni Sestorain Saralegui \authorspace Tal Schuster \\
\AND
William W. Cohen \authorspace Michael Collins \authorspace Dipanjan Das \authorspace Donald Metzler \authorspace Slav Petrov \authorspace Kellie Webster$^{\dagger}$ \\ {Google Research}}

\begin{document}

\maketitle

\begin{abstract}
Large language models (LLMs) have shown impressive results while requiring little or no direct supervision. Further, there is mounting evidence that LLMs may have potential in information-seeking scenarios. We believe the ability of an LLM to attribute the text that it generates is likely to be crucial in this setting. We formulate and study Attributed QA as a key first step in the development of attributed LLMs. We propose a reproducible evaluation framework for the task and benchmark a broad set of architectures. We take human annotations as a gold standard and show that a correlated automatic metric is suitable for development.\footnote{We publicly release all system responses and their human and automatic ratings, at \url{https://github.com/google-research-datasets/Attributed-QA}} Our experimental work gives concrete answers to two key questions ({\em How to measure attribution?}, and {\em How well do current state-of-the-art methods perform on attribution?}), and give some hints as to how to address a third  ({\em How to build LLMs with attribution?}).
\end{abstract}

\blfootnote{
$^*$ Equal contribution.
}

\blfootnote{
$^{\dagger}$ Final author.
}

\section{Introduction}

\begin{figure}[h!]
\begin{small}
\begin{tabular}{p{1.5cm}p{5cm}} 
 \hline
 \multicolumn{2}{c}{\textbf{System Input}} % align: l,c,r 
 \\
\textbf{Question}: & what is the order of the netflix marvel shows?\\ 
\\
\hline
\multicolumn{2}{c}{\textbf{System Output}} % align: l,c,r 
 \\
\textbf{Answer}: & Daredevil, Jessica Jones, Luke Cage, Iron Fist, The Defenders, The Punisher\\ 
\textbf{Attribution}: & (URL = {
% \smaller
A deal between Marvel and Netflix to produce several interconnected series was announced in November 2013, with the individual series Daredevil (2015–2018), Jessica Jones (2015–2019), Luke Cage (2016–2018), and Iron Fist (2017–2018) culminating in the crossover miniseries The Defenders (2017). A spin-off from Daredevil, The Punisher (2017–2019), was ordered in April 2016. The series were all filmed in New York State, forming the state's largest television production commitment with 161 episodes between them. [\url{https://en.wikipedia.org/wiki/Marvel's_Netflix_television_series}]}\\
 \hline
\end{tabular}
\end{small}
\caption{In attributed question answering the input to the model is a question, and the output from the model is an answer string together with a pointer to a short segment of text that supports that answer.}
\label{fig:attributedqa}
\end{figure}

Large language models (LLMs) have shown impressive results across a variety of natural language understanding and generation tasks \cite{devlin-etal-2019-bert,raffel-et-al-2020-t5,brown-et-al-2020-gpt3,rae-et-al-2021-gopher,zhang-et-al-2022,chowdhery-et-al-2022,flan-upalm} while requiring little or no direct supervision,\footnote{By “direct supervision” we refer to labeled examples for the specific task in mind, for example datasets such as the Natural Questions corpus \cite{kwiatkowski-etal-2019-natural} for question answering. We use the term “direct supervision” to distinguish this form of supervision from the term “self supervision” sometimes used in the context of LLMs.} instead using few-shot \cite{brown-et-al-2020-gpt3} or in-context learning \cite{incontext}. There is increasing evidence that LLMs  may have potential in \textit{information-seeking} scenarios, producing compelling output in scenarios ranging from “simple” question answering (e.g., \citet{kwiatkowski-etal-2019-natural,squad,joshi-etal-2017-triviaqa}), to long-form question answering \cite{multi-faceted,asqa}, and information-seeking dialog \cite{lamda, sparrow, blenderbot, webgpt}. This lack of direct supervision is particularly appealing given the difficulties of constructing labeled datasets for even simple question answering,\footnote{Here we are referring to the traditional approach to data collection for supervised learning, where human raters provide labeled examples. An alternative approach is to use an LLM to generate labeled examples that are then rated by humans. For many tasks, this latter approach is considerably simpler.} let alone more complex (but important) tasks such as multi-faceted question answering or interactive information-seeking dialog.

In many information-seeking scenarios, the ability of an LLM to attribute the text that it generates is likely to be crucial for both system developers and users (see \citet{metzler-rethinking-search, rashkin-et-al-2021, gophercite, lamda}, and section~\ref{sec:task}, for a discussion). Ideally, an “attributed LLM” would seamlessly provide evidence snippets that support the text that it generates where appropriate (specifically, whenever it makes statements about the world, e.g., see \citet{rashkin-et-al-2021}). While there has been important work in the direction of adding attribution to LLMs (see Section~\ref{sec:related}), we argue that we as a field currently have very limited understanding of the challenge and how to make progress. Critical questions are:

\begin{enumerate}
    \item How to measure attribution? 
    \item How well do current state-of-the-art methods perform on attribution? Even for the simplest possible information-seeking scenario, simple QA, this is not well understood.
    \item How to build LLMs with attribution? 
\end{enumerate}

To explore these questions, we propose Attributed Question Answering (QA). In our formulation, the input to the model/system is a {\bf question}, and the output is an ({\bf answer}, {\bf attribution}) pair where {\bf answer} is an answer string, and {\bf attribution} is a pointer into a fixed corpus, e.g. of paragraphs. The returned attribution should give supporting evidence for the answer; for example, it should satisfy the conditions in \citet{rashkin-et-al-2021} (see Section~\ref{sec:task}). Figure~\ref{fig:attributedqa} gives an example. 

\commentout{
\begin{figure}[h!]
\begin{small}
\begin{tabular}{p{1.5cm}p{5cm}} 
 \hline
 \multicolumn{2}{c}{\textbf{input}} % align: l,c,r 
 \\
\textbf{question:} & Where is the world’s largest ice sheet located today?\\ 
\\
\multicolumn{2}{c}{\textbf{output}} % align: l,c,r 
 \\
\textbf{answer:} & Antarctica\\ 
\textbf{attribution:} & (URL = {
% \smaller
\url{https://en.wikipedia.org/wiki/Ice_sheet}}, Text = The Antarctic ice sheet is the largest single mass of ice on Earth . It covers an area of almost 14 million km and contains 30 million km of ice . Around 90\% of the Earth's ice mass is in Antarctica , which , if melted , would cause sea levels to rise by 58 meters . The continent - wide average surface temperature trend of Antarctica is positive and significant at > 0.05 ° C / decade since 1957 .)\\ 
 \hline
\end{tabular}
\end{small}
\caption{In attributed question answering the input to the model is a question, and the output from the model is an answer string together with a pointer to a short segment of text that supports that answer.}
\label{fig:attributedqa}
\end{figure}}

Our motivation for studying attribution in QA is two-fold. First, it is perhaps the simplest information-seeking application, and as such it is more straightforward to evaluate. However, in spite of its simplicity, models and experiments for attributed QA are likely to be highly informative to the general goal of building attributed LLMs (see Section~\ref{sec:task} for more discussion). Second, Attributed QA is an interesting task in its own right. It has advantages over existing approaches to evaluation of question answering systems (see Section~\ref{sec:task} and Section~\ref{sec:expts}). Attribution provided by a QA system is likely to be of benefit to both system developers and users. With this motivation, we make the following contributions.

First, we define \textbf{a reproducible evaluation framework} for Attributed QA, using human annotations as a gold standard. To facilitate progress, we additionally study AutoAIS \cite{rarr}, an automatic metric that formulates evaluation as a Natural Language Inference task \cite{dagan2005pascal, snli}. We find strong correlation between the two, making AutoAIS a suitable evaluation strategy in development settings.

Further, we perform \textbf{a systematic analysis of a broad set of systems} based on state-of-the-art components, exploring different architectures and levels of supervision. While retrieve-then-read architectures are attractive for their strong performance, they typically require a large amount of data to train and can be resource intensive. We are excited by the possibility of post-hoc attribution of LLM-generated answers (though this remains challenging), and end-to-end modeling that makes limited use of QA examples. We release scored system outputs to foster further exploration, at \url{https://github.com/google-research-datasets/Attributed-QA}.

As such, our contributions give some concrete answers to questions 1 and 2 above (How to measure attribution?, and How well do current state-of-the-art methods perform on attribution?), and give some hints as to how to address question 3 (How to build LLMs with attribution?).

\section{Related Work}

\label{sec:related}

This section focuses on a few areas of related work.

\subsection{Question Answering Tasks}

Question answering has emerged as a key way to discover and demonstrate advances in LLMs.
\textbf{Reading comprehension} asks a model to take as input a question and a passage which possibly contains an answer to the question, and to extract that answer. Since the seminal work of SQuAD \cite{squad}, there has been a proliferation of reading comprehension datasets developed to benchmark different machine capabilities that are important for QA \cite{joshi-etal-2017-triviaqa,choi-etal-2018-quac,reddy-etal-2019-coqa,quizbowl}.

The Natural Questions \cite{kwiatkowski-etal-2019-natural} effort provided a large reading comprehension dataset based on real information-seeking queries to the Google\footnote{\url{https://www.google.com}} search engine, and has served more recently via Open-NQ \cite{lee-etal-2019-latent} as a benchmark for \textbf{open-domain} QA \cite{voorhees-tice-2000-trec,yang-etal-2015-wikiqa}. In open-domain QA, a system receives only an input query and must return an answer based on a set corpus of evidence.
Open-domain QA was first approached using \textit{retrieve-then-read} pipelines, which use a trained retrieval engine to identify relevant passages, before performing reading comprehension over these to deduce an answer \cite{https://doi.org/10.48550/arxiv.1704.00051}. Both retrieval and reading comprehension have been actively investigated, e.g. using neural indexing \cite{dsi,https://doi.org/10.48550/arxiv.2206.02743}, dual encoders \cite{gtr} and few-shot prompting \cite{chowdhery-et-al-2022}.
Retrieve-then-read architectures are proposed as one class suitable for Attributed QA in Section~\ref{sec:models}.
Concurrently, \textit{dense} methods that jointly optimize for passage retrieval and answer prediction \cite{lee-etal-2019-latent,dpr} have been successful, typically with less training signal than the pipeline approaches.

\citet{roberts-etal-2020} shows that T5 \cite{raffel-et-al-2020-t5} can perform a new task formulation, \textbf{closed-book} QA.
Concretely, T5 can produce answers to questions without access to any corpus at inference time, instead producing answers based on its model parameters, tuned to ``remember'' information digested in pretraining.
This result is tantalizing because it opens the possibility of more powerful question answering than so far realized in proposed task datasets.
However, it requires us to fundamentally rethink how we approach question answering and its evaluation. We defer discussion of the advantages and disadvantages of the closed-book setting to the next section.

One result of the Natural Questions dataset is that we have a subset of examples implicitly gold-labeled for attribution. NQ produced examples of the form $(x, a, c)$, where $x$ is a question, $a$ is short answer, and $c$ is a long answer (typically a paragraph) selected by annotators as support for the answer $a$. However, for a given question only a single long answer $c$ is annotated, so the set of attributed answers may be a small subset of those available on Wikipedia (the corpus also considered in our experiments). Extending from this, \citet{petroni-etal-2021-kilt} describe \textbf{the KILT benchmark} for knowledge intensive tasks (including question answering), where gold-labeled “provenance” paragraphs are provided. \citet{petroni-etal-2021-kilt} extends the NQ corpus’s coverage of provenance by using Amazon Turk annotators to mark additional paragraphs that support a given answer. The result is an increase from 1 provenance passage per (question, answer) pair to an average of 1.57 passages.

\subsection{LLMs with Attribution} 

Existing work has explored whether attribution may be achieved using retrieval. \citet{rarr} proposes a two-stage technique where LLM-generated text is post-edited to be made attributable to web content retrieved in a first stage. \citet{gophercite} propose a system, GopherCite, and perform a close set of evaluations to ours. In GopherCite, an LLM generates answers to questions using evidence retrieved by Google search for the given query as its input, along with paragraph-level supporting evidence that is evaluated using human raters. The guidelines are not specified in precise detail, though appear similar to AIS. 

GopherCite is an important reference, but we see two limitations. First, only a single system, the hybrid Google search/LLM system, is evaluated. Second, the evaluation is limited. Of 307 questions presented to raters, only 115 questions were retained for evaluation, with the remaining 192 (62.5\% of questions) being discarded due to raters skipping some items. Little is said about the basis on which raters skipped items, but this makes the 80\% accuracy of the system hard to interpret.

There has been a recent flurry of activity in producing compelling proof-of-concept demos that generate seemingly factual responses in information-seeking dialog settings \cite{webgpt,sparrow,lamda,realm}. These operate by incorporating a retrieval system, typically a commercial search engine such as Bing,\footnote{\url{https://www.bing.com/}} into an LLM that then conditions its output on the retrieved content. The promise of these demos is a key motivation for this work: we perform a systematic study of different architecture decisions, using the principles in the AIS work \cite{rashkin-et-al-2021}: recent demos most closely fit into the \textit{Retrieve-then-read} class of architectures in Section~\ref{sec:models}, where other possible design choices are described, and each operationalize the concept of attributability differently.

\section{Attributed Question Answering}

This section defines the Attributed QA task, and gives discussion.

\subsection{Task Definition} 

\label{sec:task}

We assume a set ${\cal C}$, which is a fixed set of units to which answers can be attributed. For example, ${\cal C}$ might be the set of all paragraphs in some corpus. More specifically, each $c \in {\cal C}$ is the ID for some unit; we use $\hbox{\tt text}(c)$ to refer to the actual paragraph text for natural language datasets.
The input to an attributed QA system $g$ is a question $x$. The output from the system is a pair $g(x) = (a, c)$, where $a$ is a text string, and $c$ is a member of ${\cal C}$. 

\subsection{Evaluation}

We consider two evaluation metrics for the Attributed QA task: first, human ratings that are the gold-standard, and second, automatic evaluation methods, which we show can be suitable in development settings. Section~\ref{sec:expts} gives analysis of the correlation between the two.

\paragraph{Human Evaluation.} Given a triple $(x, a, c)$, we use the AIS evaluation definitions and guidelines  \cite{rashkin-et-al-2021}  to judge whether the answer to question $x$ is attributable to $c$. Raters are asked to answer the following two questions, in the context of the question $x$ (where the system response is the answer $a$, and the source document is $c$):
\begin{enumerate}
    \item Is all of the information relayed by the system response $(a, c)$ interpretable to you?
\item Is all of the information provided by the system response $a$ fully supported by the source document $c$? 
\end{enumerate}
We define the rating of $(x, a, c)$ as “attributable” if the answer to both of these questions is “yes”.

Assume a set of test questions $x_1 \ldots x_n$, and a system $g$ to be evaluated. Define $r_i$ to be the (randomly chosen) pool of raters on the $i$th test example, and $h(x_i, g(x_i), r_i)$ to be $1$ if the majority of the annotators mark the system output $g(x_i)$ to be attributable, or $0$. The test accuracy is then 
\[
E[g] = \frac{1}{n} \sum_{i=1}^n h(x_i, g(x_i), r_i)
\]
That is, the test accuracy is simply the proportion of test examples where the majority of the raters judge the system’s output to be attributable.\footnote{This is an estimate of $E\left [ h(X, g(X), R) \right ]$ where the expectation is taken over the random choice of example $X$, and the random choice of raters $R$.}

\paragraph{Automatic Evaluation (AutoAIS).} In addition, we will make extensive use of an automatic measure, based on the NLI classifier of \citet{honovich-etal-2022-true-evaluating}, AutoAIS \cite{rarr}. See Section~\ref{sec:expts} for full details of the classifier. Taking $\hbox{\tt AutoAIS}(x_i, g(x_i))$ to be the output of the NLI classifier ($1$ for attributable vs $0$ for non-attributable), we define
\[
E^{\hbox{\tt A}}[g] = \frac{1}{n} \sum_{i=1}^n \hbox{{\tt AutoAIS}}(x_i, g(x_i))
\]

\subsection{Discussion}

Given this definition, we make the following remarks concerning the complexity of the task, motivation for the task, and the relationship to the more general attributed LLM problem.

\paragraph{Remark 1: Size of the Label Space.} We note that for many queries, the set of labels $(a, c)$ that form a correctly attributed answer to $x$ is likely to be large. This is due to two causes. First, for a given answer $a$, there may be multiple paragraphs $c \in {\cal C}$ that support that answer. Second, for many queries, there may be a diverse set of answers that have some supporting paragraph in ${\cal C}$. This diversity comes from several sources, for example: the same underlying answer being expressed by different strings; differing opinions about the answer to a question; differing answers under differing interpretations of a query. This makes evaluation of QA systems—whether or not attributed—challenging.

\commentout{
\begin{figure}
    \centering
\begin{tabular}{|l|l|}
{\bf Question}&{\bf Multiple attributable answers}
where was the first colony in north america located & Veracruz, Greenland \\
     & 
\end{tabular}
    \caption{Some examples from our data where the same question has multiple attributable answers.}
    \label{fig:multiple-examples}
\end{figure}}

\commentout{(Examples: include figure showing examples of differing answers to the same question. Chinese GDP answer; 1953 vs. June 1953; small intestine vs. ilieum; Roanoke vs. Greenland; etc Caption: Sources of diversity include: the same underlying answer may be expressible using different strings; there may be differing opinions about the answer to a question; there may be differing answers under differing interpretations of a query.)}

\paragraph{Remark 2: Motivation for Attribution.} \citet{rashkin-et-al-2021, lamda,  gophercite} give extensive motivation for attribution in LLMs. We focus on a few key points here.
First, attribution allows either a system developer or user to see the underlying source supporting an answer, and to assess aspects including trustworthiness and nuance. As such, attribution deliberately avoids the need for judgments of the “factuality” of claims, something that is challenging for all but the most simple questions, see \citet{rashkin-et-al-2021}.

Second, attribution offers system developers a more streamlined human evaluation of answer quality. Consider instead QA definitions where a model simply outputs an answer string. Evaluation of new answer strings would require either:
\begin{enumerate}
    \item Humans to use a search engine to attempt to find evidence for the answer. This places significant onus on raters and may be intractable;
    \item The curation of one or more gold labels $y_i$ for each test example $x_i$, together with test error defined as $1/n \sum_{i=1}^n L(g(x_i), y_i)$ for some similarity measure $L$ over answer strings. This is the closed-book QA setting.
\end{enumerate}

\paragraph{Remark 3: Comparison to Closed-Book QA Evals.} The closed book QA setting has been valuable in developing LLMs and QA systems \cite{roberts-etal-2020}, and provides a highly effective and convenient measure of LLM performance. They do, however, have two significant drawbacks:
\begin{enumerate}
    \item Closed-book QA evals do not require a system to provide attribution for its answers. In this sense closed-book evals measure performance on a task that is arguably incomplete or of limited utility to users and system designers.
    \item Closed-book QA evals depend on gold-curated labels $y_i$ for test examples, leading to significant difficulties for questions with diverse answers. In this sense there is a risk that closed-book evals significantly undercount performance (but without large-scale human evaluations such as those described in the current paper, it is impossible to estimate the scale of this problem).
    \end{enumerate}

\paragraph{Remark 4: Motivation for Human Ratings.} Throughout this paper we will take the final measure of system performance to be based on human ratings. A primary motivation for this is that given the size of the space of attributable $(a, c)$ pairs (Remark 1), it is unclear whether curating gold standard $(a, c)$ labels that cover enough of the output space is feasible for the task: or at least if attempts are made to do this, we will need to correlate with human evals to measure their effectiveness.

One side-effect of the large quantity of human labels gathered in this paper is that we may be in a much better position to develop high quality automatic evals for attributed QA. For example, we can measure the level of correlation between existing measures such as AutoAIS and human ratings. Or we can use the labels gathered to train automatic evals (see \citet{sellam-etal-2020-bleurt, bulian2022tomayto}), potentially with a large set of reference (answer, attribution) pairs for each dev/test example.

\paragraph{Remark 5: Relationship of Attributed QA to Attributed LLMs.} Attributed QA is perhaps the simplest possible attributed LLM task, but it gets at the core task of attribution of “statements” or “propositions”: see \citet{rashkin-et-al-2021} for definitions and discussion. In short, the problem of attributing a question/answer pair (e.g., $x =$ “when did the first dinosaurs live”, $a =$ “230 million years ago”) is closely related to the problem of attributing statements made by an LLM, where a “statement” is some declarative sentence, for example “the first dinosaurs lived 230 million years ago”. Much of LLMs’ outputs in more complex information-seeking scenarios such as dialog and multi-faceted QA involve sequences of such statements (or, essentially equivalent, answers to questions), many of which require attribution. There are undoubtedly complexities in extending results for attributed QA to the full attributed LLM problem—for example deciding which statements need to be attributed, dealing with the move from question/answer pairs to more general statements, or dealing with complex statements that may involve attribution to multiple sources. But our working hypothesis is that progress on Attributed QA will extend naturally into more complex tasks.

\section{Approaches to Attributed QA}

\label{sec:models}

We now describe the different systems investigated in this paper. At a high level, they fall into the following three architecture classes, and may be differentiated in terms of the type and quantity of supervision that is used. We will study a variety of different systems that fall into these three categories, carrying out ablations of key components. 

\subsection{Architectures}

\paragraph{Retrieve-then-read (RTR).} Following approaches to open-domain QA, retrieve-then-read (RTR) models first perform retrieval of $k$ relevant passages based on the input question alone, where $k$ is a relatively small number. A second-stage model then takes $P \subset k$ retrieved passages, possibly reranked, as input to generate a short answer, and chooses one of $A \subset k$ retrieved passages as support for that answer. 

In our experiments, we used BM25 \cite{robertson2009probabilistic} for sparse retrieval, GTR \cite{gtr} for dense retrieval, and Fusion-in-Decoding \cite[FID,][]{izacard-etal-2022} for answer generation. FID may be trained with $T \subset k$ retrieved passages as input to answer generation, to reduce memory requirements. GTR may be used in the open-source version (PT-GTR in following tables), or further tuned for NQ (GTR).

\paragraph{Post-hoc retrieval.} In these systems, an LLM is first used to generate an answer to the input question, typically using few-shot prompting, without any use of retrieval. The question and answer are then concatenated to form a query to sparse or dense retrieval, again giving $k$ relevant passages. For $k > 1$, a final step selects the highest scoring passage containing the answer generated by the LLM as the attribution. 

\paragraph{LLM-as-retriever.} In LLM-as-retriever models \cite{dsi,https://doi.org/10.48550/arxiv.2206.02743}, an LLM is used to generate both an answer and a pointer into the attribution corpus through some combination of prompting and fine-tuning but without any use of either sparse or dense retrieval. In this paper, we split attribution into a two-stage process, with the LLM first generating a webpage URL, from which a paragraph is then selected as support for the answer. A natural extension of this approach (not investigated in here) would be to generate a pointer to a paragraph rather than a URL.

\subsection{Supervision}

A second important axis in which systems can be differentiated concerns the type and quantity of supervision that is used. In {\bf NQ-64} systems, very limited supervision, in the form of 64 randomly chosen training examples from the Natural Questions (consisting of question/answer pairs) is used. In {\bf NQ-full} systems, we assume access to the full NQ training set. In RTR pipelines, GTR retrieval and FID answer generation use NQ-full. When NQ-full is used to select exemplars in post-hoc pipelines, the 64 most similar examples to the target based on the BM25 score are used.

%\paragraph{Fine-tuning or prompting.} {\bf NQ-64} or {\bf NQ-full} data can be used as supervision either through fine-tuning some underlying model, or through few-shot prompting. We will consider both of these approaches for various components of the models.

\subsection{Best Systems}

The next section of the paper describes experiments with a number of systems. We briefly highlight four particularly important ones (and two variants) that achieve the highest AIS score for their architecture.

\paragraph{Best RTR system.} GTR is first used for retrieval of top $k=50$ passages from an input query, and the NQ-reranker is then used to rerank these passages. FID is trained with $T = 50$ passages but generates an answer based on the top $P=1$ passage, which is returned as the attribution. (Note that this approach is very close to the approach of \citet{izacard-etal-2022}, with the added final passage selection step that allows evaluation for attribution.)

\paragraph{Best post-hoc retrieval system.} A prompted version of a 540B parameter PaLM produces an answer to the question. The prompts are 64 question/answer pairs from the Natural Questions training set, chosen based on BM25 similarity. GTR is then used for retrieval of an attribution, using the question concatenated with answer as the input query, by selecting the passage in the top $k$ = 50 which contains the PaLM-predicted answer string.

\paragraph{Best low-resource system.} A prompted version of a 540B parameter PaLM produces an answer to the question. Again, prompts are 64 question/answer pairs selected from the full Natural Questions training set. BM25 is used for post-hoc retrieval, again with the question/answer pair concatenated to form the query. We refer to this system as “very close to unsupervised” as it only requires 64 NQ examples, it does not require fine-tuning, and the underlying retrieval method does not require supervision or fine-tuning.

\paragraph{Best LLM-as-retriever system.} We explore the possibility of more end-to-end approaches to Attributed QA by fine-tuning a 540B parameter PaLM to generate an answer and Wikipedia URL, given a question input. The fine-tuning data used for this was questions generated by a question-generation model trained on SQuAD \cite{squad} over decontextualized \cite{choi-etal-2021-decontextualization} Wikipedia sentences, as well as the Natural Questions training set. The Wikipedia paragraph with the highest BM25 score is used as the attribution.

\paragraph{AutoAIS reranked variants.} To fairly assess how good RTR and PaLM post-hoc pipelines are at producing answers which \textit{could} be attributed, we additionally experiment with system variants where AutoAIS is used as a reranker. These are identical to the above RTR and post-hoc systems, but instead use AutoAIS scores to select the attribution passages: the retrieved passage in the top $k = 50$ with highest AutoAIS score is selected as the attribution (and so is used to generate the answer in RTR). Since these variants use AutoAIS as a system component, we evaluate performance only on AIS (not AutoAIS). We encourage others who make such use of automatic evaluations to improve system quality to similarly distinguish between when they are being used as system components and when they are being used for evaluation.

\section{Experiments}

\label{sec:expts}

We now describe experiments on Attributed QA. We first give technical details, then present system results, before concluding with an analysis of the evaluation metrics.

\subsection{Datasets}

\paragraph{Question Set.}

We evaluate the short-answer seeking questions from the validation set of the Natural Questions \cite{kwiatkowski-etal-2019-natural}, i.e. those that appear in OpenNQ \cite{lee-etal-2019-latent}. 

\paragraph{Attribution Corpus.} We use a snapshot of Wikipedia from 2021-10-13 to derive ${\cal C}$, using Pyserini\footnote{\url{https://pypi.org/project/pyserini/}} to extract paragraphs from each page.

\subsection{Evaluation Metrics}

We report three metrics for all experiments. 

\paragraph{(Human) AIS} The gold-standard metric is the AIS measure assessed by human raters, as described in Section~\ref{sec:task}. Raters are trained using repeated annotations with feedback, until reaching high performance on the task and we take the majority vote from 5 raters. Given the cost of human rating, we evaluate on 1000 randomly-chosen questions and estimate standard errors using two-sided bootstrap re-sampling\footnote{\url{https://docs.scipy.org/doc/scipy/reference/generated/scipy.stats.bootstrap.html}}.

\paragraph{AutoAIS} AutoAIS formulates evaluation as a Natural Language Inference task that asks a model whether the question and answer are entailed by the provided attribution. We use a T5 \cite{raffel-et-al-2020-t5} checkpoint with 11B parameters fine-tuned on a collection of NLI-related tasks \citep{MNLI,snli,thorne-etal-2018-fever,zhang-etal-2019-paws,DBLP:conf/aaai/KhotSC18,schuster-etal-2021-get}. We score a given (premise, hypothesis) input by measuring the output probability when force-decoding the positive label, resulting in a score between 0 (no entailment) and 1 (entailment). We treat values $\geq 0.5$ as indicating valid (answer, attribution) predictions.

\paragraph{Exact Match (EM)} Finally, for comparison to prior work, we also report EM\footnote{\url{https://github.com/google-research/text-to-text-transfer-transformer/blob/2ce1574a0c2f5ed65a08e87cc38ad8ceb222b239/t5/evaluation/metrics.py\#L154}} for the answer string alone, ignoring attribution.

\subsection{System Results}

\begin{table}
\centering
\begin{small}
% https://colab.corp.google.com/drive/11EW-765Mm7bmXwjOZNIeYCuloNIfjw-P?resourcekey=0-Z35VJ_oC2z08Gb92y_0xPg&userstoinvite=websterk%40google.com&actionButton=1#scrollTo=KTnRAH-edjm6
\begin{tabular}{|l||ccc|}
\hline
        Architecture &    EM & AutoAIS &             AIS \\
\hline
\hline
  Retrieve-then-read &  41.1 &    66.3 &  $65.5 \pm 1.5$ \\
 + AutoAIS reranking &  53.3 &       - &  $71.4 \pm 1.4$ \\[5pt]
  Post-hoc-retrieval &  49.5 &    53.9 &  $55.6 \pm 1.5$ \\
 + AutoAIS reranking &  49.5 &       - &  $59.0 \pm 1.5$ \\[5pt]
        Low resource &  39.5 &    41.9 &  $48.6 \pm 1.6$ \\
    LLM-as-retriever &  50.1 &    41.5 &  $46.0 \pm 1.6$ \\
\hline
\end{tabular}
\end{small}
\caption{\footnotesize{Results for the highest-AIS systems in each architecture and reranked variants, as outlined in Section~\ref{sec:models}. 
With AutoAIS reranking, AutoAIS is used to select attribution passages, to assess how good the system is at producing answers which \textit{could} be attributed. AutoAIS is not reported for reranked variants given its use as a system component.}}
\label{fig:bestresults}
\end{table}

Table~\ref{fig:bestresults} shows results for the systems in each architecture class with the best AIS score, with AutoAIS Reranked variants. The most striking result is that \textbf{\textit{the systems which perform best on AIS do not necessarily achieve the strongest EM accuracy}} (cf. Tables~\ref{fig:rar-results} and \ref{fig:posthoc-results}). This is discussed below in Section~\ref{sec:correlation}, where we find EM correlates only modestly with human judgment of AIS and has important limitations for Attributed QA evaluation. At the same time, we note that we did no special modeling to maximize EM score, such as instruction tuning \cite{flan} or chain of thought prompting \cite{chainofthought}, and that models tuned for greater EM may also achieve higher AIS scores.

\textbf{Best RTR \textit{achieves the highest performance}} ($p \ll 10^{-5}, t=4.55$, in comparison with the best non-RTR system), despite using LLMs with relatively small numbers of parameters (using T5 XL with 3B parameters, compared to PaLM with 540B). However, RTR approaches have the shortcoming that they require relatively large amounts of explicit supervision, for example in the form of NQ examples (an open question is whether RTR systems with much less supervision can be developed). They are also likely to be highly dependent on the accuracy of the retrieval step.

It is encouraging that \textbf{Best Post-hoc} achieves relatively high EM because it requires minimal amounts of supervision for answer generation (using prompting). However, these models generally require LLMs with large numbers of parameters\footnote{For example, \citet{chowdhery-et-al-2022} appendix H.1 reports NQ exact match results of 14.6/27.6/39.6\% for 8/64/540 billion parameter models, showing significant performance increases with increasing scale.} (presumably needed for memorization). Also, attribution poses a challenge in this setting; as noted above, on AIS the best RTR system is significantly better than the best post-hoc system, and this difference carries over to AutoAIS as well. However, since reranking \textit{is} able to find good attribution passages, this result suggests that \textbf{\textit{attribution is more difficult in a post-hoc setting than in RTR}}, and is a key area for future development.

\textbf{Best Low Resource} performs competitively with Best Post-hoc on AIS and AutoAIS despite using a sparse retrieval. This is promising for more complex information-seeking tasks where it is challenging to provide explicit supervision, and where LLMs have been shows to provide fluent output.

End-to-end models have the potential benefit of not requiring retrieval at all. That the performance of \textbf{Best LLM-as-retriever} is competitive with low-resource post-hoc attribution is promising, given that it is BM25 which is used to select a paragraph from the returned URL. However, they again require LLMs with large numbers of parameters. 

\begin{table*}
\centering
\begin{small}
% https://colab.corp.google.com/drive/11EW-765Mm7bmXwjOZNIeYCuloNIfjw-P?resourcekey=0-Z35VJ_oC2z08Gb92y_0xPg&userstoinvite=websterk%40google.com&actionButton=1#scrollTo=87kA4jXvUs27
\begin{tabular}{|l||l|lll||c|cc|}
\hline
 System & Retrieval &   T &   P &   A &    EM & AutoAIS &             AIS \\
\hline
\hline
  RTR-1 &      BM25 &   1 &   1 &   1 &  27.7 &    16.6 &               - \\
  RTR-2 &      BM25 &  50 &   1 &   1 &  20.2 &    23.7 &  $26.0 \pm 1.4$ \\
  RTR-3 &      BM25 &  50 &  50 &   1 &  45.6 &    16.1 &               - \\
  RTR-4 &      BM25 &  50 &  50 &  50 &  45.6 &    42.9 &  $48.5 \pm 1.6$ \\[5pt]
  RTR-5 &    PT\_GTR &   1 &   1 &   1 &  40.0 &   47.2 &               - \\
  RTR-6 &    PT\_GTR &  50 &   1 &   1 &  38.9 &   53.2 &  $53.8 \pm 1.6$ \\
  RTR-7 &    PT\_GTR &  50 &  50 &   1 &  52.9 &   41.9 &               - \\
  RTR-8 &    PT\_GTR &  50 &  50 &  50 &  52.9 &   59.3 &  $60.0 \pm 1.5$ \\[5pt]
  RTR-9 &       GTR &   1 &   1 &   1 &  46.0 &    58.8 &  $58.7 \pm 1.6$ \\
 RTR-10 &       GTR &  50 &   1 &   1 &  41.1 &    66.3 &  $65.5 \pm 1.5$ \\
 RTR-11 &       GTR &  50 &  50 &   1 &  53.3 &    50.1 &  $51.0 \pm 1.6$ \\
 RTR-12 &       GTR &  50 &  50 &  50 &  53.3 &    64.1 &  $63.3 \pm 1.5$ \\
\hline
\end{tabular}

\end{small}
\caption{\footnotesize{Ablations for Retrieve-then-read (RTR) systems. {\bf T} Number of passages used for training FID. {\bf P} the number of retrieved passages input to answer generation. {\bf A} = "1" if the top $1$ retrieved passage was returned as attribution, "50" if the passage scored highest by the retrieval system was chosen from the top 50, under the constraint that the answer string was in the passage.}}
\label{fig:rar-results}
\end{table*}

\subsection{Ablations}
Ablation studies are presented for RTR systems in Table~\ref{fig:rar-results} and for post-hoc retrieval systems in Table~\ref{fig:posthoc-results}. In the RTR models, the best dense-retrieval system (RTR-10) outperforms the best sparse-retrieval system (RTR-4) by 17 points AIS ($p \ll 10^{-13}, t=7.79)$. Among the post-hoc systems, the dense retrievers also have the edge with the AIS difference between the best systems of each class (Post-6 vs Post-2) being statistically significant ($p \ll 0.01, t=2.91$).

For RTR systems, training FID with $T = 50$ examples seems essential for achieving top performance, though using $P = 50$ passages for answer generation is only useful if all $A = 50$ passages are considered for attribution also. Simply selecting the top retrieved passage ($A = 1$) as the attribution after training and generating an answer with 50 passages performs poorly (e.g., $p \ll 10^{-7}, t=5.60$ for RTR-12 vs RTR-11 on AIS). That is, while the best performing architecture, \textbf{\textit{RTR is resource intensive and it is unclear how to reduce this without hurting performance}}.

Across the post-hoc systems, selecting an attribution passage among the $k = 50$ that are retrieved seems better than simply using the top $k = 1$ (e.g., $p\ll.01, t=2.78$ for Post-5 vs Post-6 but $p=0.04, t=2.01$ for Post-1 vs Post-2). The interesting trend here is in the impact of expanding the pool of NQ examples for exemplar selection. Using NQ-full gives a 10 point boost to Exact Match for both BM25 and GTR systems but the impact on AIS is much smaller.

\begin{table*}
\centering
\begin{small}
% Eval post hoc

\begin{tabular}{|l||l|ll||c|cc|}
\hline
 System & Retrieval & Exemplars &  $k$ &    EM & AutoAIS &             AIS \\
\hline
\hline
 Post-1 &      BM25 &   NQ-full &    1 &  49.5 &    42.8 &  $47.8 \pm 1.6$ \\
 Post-2 &      BM25 &   NQ-full &   50 &  49.5 &    45.3 &  $49.1 \pm 1.6$ \\
 Post-3 &      BM25 &   NQ-64   &    1 &  39.5 &    39.9 &  $46.9 \pm 1.6$ \\
 Post-4 &      BM25 &   NQ-64   &   50 &  39.5 &    41.9 &  $48.6 \pm 1.6$ \\[5pt]
 Post-5 &       GTR &   NQ-full &    1 &  49.5 &    48.5 &  $49.4 \pm 1.6$ \\
 Post-6 &       GTR &   NQ-full &   50 &  49.5 &    53.9 &  $55.6 \pm 1.5$ \\
 Post-7 &       GTR &   NQ-64   &    1 &  39.5 &    44.2 &  $47.4 \pm 1.6$ \\
 Post-8 &       GTR &   NQ-64   &   50 &  39.5 &    50.1 &  $51.9 \pm 1.6$ \\
\hline
\end{tabular}

\end{small}
\caption{\footnotesize{Ablations for post-hoc retrieval systems. {\bf Exemplars} = number of exemplars used in the PaLM prompt: "NQ-64" means 64 Natural Questions examples were chosen at random, "NQ-full" means that 64 NQ examples were chosen based on a BM25-defined distance measure.}}
\label{fig:posthoc-results}
\end{table*}

Taken together, these results show that existing state-of-the-art methods are suitable for Attributed QA, though there is still headroom to improve, especially in the post-hoc attribution of LLM-generated answers. As to how to design systems, we have discussed how it depends on many factors which should be carefully considered.

\subsection{Correlation between AIS and EM/AutoAIS}
\label{sec:correlation}

We now focus on the question of how to best measure attribution given our observations so far. To do this, we estimate the correlation between system scores on (human) AIS, and EM and AutoAIS in turn, by calculating the Pearson coefficient between the two sets of scores (i.e. between AIS and EM scores, and between AIS and AutoAIS scores). 

\paragraph{EM} We saw above that best AIS performance did not necessarily go hand-in-hand with best EM accuracy. Consistent with this, the Pearson correlation coefficient between the system EM and AIS scores is modest, at 0.45 (see Figure~\ref{fig:em-corr}). Manual analysis of the disagreements revealed multiple factors to be involved, including answers with inexact string matches to the NQ reference answer, stale reference answers, and questions with more than one valid answer able to be retrieved (see Table~\ref{tab:examples}). Overall, we suggest that \textbf{\textit{our results point to the limitation of reference answer corpora and string matching evaluation for future research}}.

\paragraph{AutoAIS} 

On the other hand, correlation between system AIS and AutoAIS scores is remarkably strong, with a Pearson coefficient of 0.96 (Figure~\ref{fig:autoais-corr}). This suggests that \textbf{\textit{AutoAIS is fit-for-purpose as a development metric at the aggregate level}} (provided it is not used as a system component). 
\begin{figure}
%\centering
\hspace{-1cm}
\begin{tikzpicture}
  \begin{axis} [xlabel=AIS,ylabel=EM,legend pos=north west,xmin=20,ymin=20,xmax=80,ymax=80]
    \input{data/ais_em_data}
  \end{axis}
\end{tikzpicture}
\caption{\footnotesize{System-level correlation between AIS and EM scores. Each mark represents a system result from the Ablation. The dashed line represents a line-of-best-fit, with Pearson correlation of 0.45.}}
\label{fig:em-corr}
\end{figure}
\begin{figure}
\centering
\begin{tikzpicture}
  \begin{axis} [xlabel=AIS,ylabel=AutoAIS,legend pos=north west,xmin=20,ymin=20,xmax=80,ymax=80]
\input{data/ais_auto_ais_data}
  \end{axis}
\end{tikzpicture}
\caption{\footnotesize{System-level correlation between AIS and AutoAIS scores. Each mark represents a system result from the Ablation. The dashed line represents the line-of-best-fit, with strong Pearson correlation of 0.96.}}
\label{fig:autoais-corr}
\end{figure}

To get a deeper understanding of the correlation, we followed up with an instance-level correlation study where the data series are the per-question ratings for a given system. Correlation was much lower and more variable here. Therefore, we recommend care should be taken against reading individual AutoAIS scores too closely. 

The reranked variants are outliers to this strong correlation, with attributions selected by AutoAIS scoring lower on human evaluation than would be expected based on a linear fit. This is consistent with instance-level AutoAIS (which was used in reranking) being noisier than system-level AutoAIS: the passage with the best AutoAIS is not necessarily the one preferred by humans. 

\section{Future Directions}

We see many exciting areas for future work.

\paragraph{Modeling.} While retrieve-then-read systems achieve strong performance, this class typically requires a large amount of data to train and can be resource intensive. We are excited by the possibility of post-hoc attribution of LLM-generated answers and end-to-end modeling for Attributed QA. Future directions to improve performance in these settings includes studying the challenge of retrieval for post-hoc attribution, and devising training signals for end-to-end modeling. One possible, albeit noisy, source for the latter is AutoAIS, which we observed correlated well at the system-level with human judgments of AIS. We also noted the promise of instruction tuning and chain of thought prompting for improving the quality of LLM-generated answers.

\paragraph{Evaluation.} We observed that AutoAIS was fit-for-purpose as a development metric, but had shortcomings including only moderate correlation with human ratings at the instance-level. There are at least two possible ways to use the human rating data collected in this paper to improve from this. First, the data could form a cache used to score system predictions which have been observed previously. In this way, the data could be seen as an extension of KILT \cite{petroni-etal-2021-kilt}, curating a range of attributed answers that do not require further verification. A softer approach could apply prior work \cite{sellam-etal-2020-bleurt,rei-etal-2020-comet} and use the data to learn an improved automatic evaluation metric for attribution. We note that the latter is additive with using AutoAIS as a noisy training signal for end-to-end learning.

\paragraph{Tasks.} We have presented an in-depth study on the Natural Questions to demonstrate the promise of an attribution task with automatic and human verification. However, our best system requires use of the full NQ training set. We would like to understand how general-purpose this approach is, and whether systems that make less use of direct supervision transfer to new settings better. Therefore, we see future work in evaluating on different datasets \cite[esp.][]{joshi-etal-2017-triviaqa}, perhaps with multilingual \cite{clark-etal-2020-tydi} or multimodal \cite{vqa} attribution. We are excited by the challenge of attributing generated text more generally, perhaps in long-form QA \cite{asqa}.

\section{Conclusion}

We establish a research agenda to develop attributed large language models. We believe that attribution will be crucial for technologies based on LLMs in information-seeking settings. To understand how to make progress in this area, we define and study a new task, Attributed QA, which bases evaluation on the AIS principles and benchmarks architecture designs using a range of state-of-the-art components to build systems. We consider human rating to be the gold standard for system evaluation, but find that AutoAIS correlates well with human judgment at the system level, offering promise as a development metric where human rating is infeasible, or even as a noisy training signal. Retrieve-then-read approaches achieve the strongest performance on our evaluation, but require full use of a traditional training set. Post-hoc attribution appears to be a viable architecture for future work, but remains challenging.

\section{Ethical Considerations}

The main ethical consideration of this work concerns "factuality." As in \cite{rashkin-et-al-2021}, we observe that it is incredibly challenging to judge whether any but the simplest claim is \textit{factual}. Instead, for most questions, there will be multiple valid answers that are distinguished by nuances that can be subtle. Therefore, we believe attribution will be crucial in most information-seeking scenarios and explore what it means for an LLM to be able to attribute text it generates. In this way, users can inspect sources to make their own judgment of trustworthiness and answer scope. It is an interesting research question not studied here, how to identify issues like factual inaccuracies and biases in web sources.

We also consider the issue that Attributed QA is only explored in English using, for the most part, resource-intensive approaches that may not be accessible to many. To encourage future work that expands from here, the AIS principles are publicly available \cite{rashkin-et-al-2021} and we have released all system outputs and their ratings. We are excited by the promise of low-resource and end-to-end solutions to meet the diverse challenge of attribution in language modeling.

\section{Contributions}

Bernd Bohnet, Vinh Q. Tran, and Pat Verga lead the technical work for this paper, including implementing models, running experiments, analyzing results and making improvements. Kellie Webster acted as TL.

Livio Baldini Soares built the core infrastructure for BM25 retrieval, Ji Ma and Jianmo Ni contributed the dense retrieval and reranking pipelines, and Kai Hui helped with PaLM usage. Daniel Andor and Kuzman Ganchev ran components that enabled the LLM-as-retriever model. 

Bernd Bohnet managed the human rating collection and its data pipeline. Roee Aharoni and Jonathan Herzig trained the NLI models for automatic evaluation and Roee open sourced the model on huggingface for public use. Massimiliano Ciaramita conceived the AutoAIS prompt and Lierni Sestorain Saralegui explored several variants. Tom Kwiatkowski, Livio Baldini Soares, and Daniel Andor built the Attribution corpus and helped with datasets. Kellie Webster implemented the standard automatic evaluation script around this work.

Michael Collins and Kellie Webster were the primary writers of the paper. William W. Cohen helped with the related work, Jacob Eisenstein contributed the statistical analysis in the results section. William W. Cohen, Michael Collins, Dipanjan Das, Don Metzler, Slav Petrov, and Kellie Webster developed the direction for this work and contributed significant feedback during paper writing. 

\section{Acknowledgements}

We would like to thank our many colleagues whose insightful discussion shaped this work, including Fernando Pereira, Ankur Parikh, Jon Clark, Marc Najork, and Vitaly Nikolaev. The human rating process was managed by Muqthar Mohammad and Isabel Kraus-Liang, who worked diligently to produce incredible results. Kathy Meier-Hellstern, Suneet Dhingra, and teams provided invaluable support.

\appendix

\section{Examples}

We identify three classes of interesting examples that demonstrated the value of AIS over EM.

\begin{table*}
    \centering
    \begin{tabular}{p{0.25\textwidth}|p{0.72\textwidth}}
    \toprule
    Inexact String Match & \textbf{Question}: \textit{who does marge's voice on the simpsons} \\
    & \textbf{Reference answer}: ['Julie Deborah Kavner'] \\
    & \textbf{System prediction}: Julie Kavner \\
    & \textbf{System attribution}: Kavner became most famous for her role as Marge Simpson on the animated television show The Simpsons, a show that continues to the present day. The Tracey Ullman Show included a series of animated shorts about the dysfunctional Simpson family. Voices were needed for the shorts, so the producers decided to ask Kavner and fellow cast member Dan Castellaneta to voice Marge and Homer rather than hire more actors. Kavner has what Hilary de Vries of The New York Times described as a "honeyed gravel voice". Kavner says her distinctive voice is due to "a bump on [her] vocal cords". Marge's voice has aged considerably throughout the series as Kavner has gotten older. [\url{https://en.wikipedia.org/wiki/Julie_Kavner}] \\
    \midrule
    Stale reference answer & \textbf{Question}: \textit{when was the last time clemson won the national championship} \\
    & \textbf{Reference answer}: ['2016'] \\
    & \textbf{System prediction}: 2018 \\
    & \textbf{System attribution}: Formed in 1896, the program has over 750 wins and three consensus national championships in the modern era. Clemson was a College Football Playoff finalist in 2015, 2016, 2018, and 2019, winning the championship game over Alabama in 2016 and 2018. Clemson has had six undefeated seasons, six consecutive playoff appearances, 26 conference championships, and eight divisional titles. Its alumni includes over 100 All-Americans, 17 Academic All-Americans and over 250 players in the National Football League. Clemson has had seven members inducted into the College Football Hall of Fame: players Banks McFadden, Terry Kinard, and Jeff Davis along with coaches John Heisman, Jess Neely, Frank Howard, and Danny Ford. [\url{https://en.wikipedia.org/wiki/Clemson_Tigers_football}] \\
    \midrule
    Multiple valid answers & \textbf{Question}: \textit{where did the tea come from in the boston tea party} \\
    & \textbf{Reference answer}: ['England', 'East India Company', 'the East India Company'] \\
    & \textbf{System prediction}: China \\
    & \textbf{System attribution}: The Boston Tea Party was an American political and mercantile protest by the Sons of Liberty in Boston, Massachusetts, on December 16, 1773. The target was the Tea Act of May 10, 1773, which allowed the British East India Company to sell tea from China in American colonies without paying taxes apart from those imposed by the Townshend Acts. The Sons of Liberty strongly opposed the taxes in the Townshend Act as a violation of their rights. Protesters, some disguised as American Indians, destroyed an entire shipment of tea sent by the East India Company. [\url{https://en.wikipedia.org/wiki/Boston_Tea_Party}] \\
    \bottomrule
    \end{tabular}
    \caption{System predictions that AIS reveals as valid but that are scored as incorrect by Exact Match. We note three key classes of errors.}
    \label{tab:examples}
\end{table*}

\end{document}